\pdfoutput=1

\documentclass[11pt]{article}

\usepackage{emnlp2021}
\usepackage{times}
\usepackage{latexsym}
\usepackage[T1]{fontenc}
\usepackage[utf8]{inputenc}
\usepackage{microtype}

\usepackage{graphicx}
\usepackage{amsfonts} 
\usepackage{amsmath}  
\usepackage{multirow} 
\usepackage{microtype} 
\usepackage{subfigure}  
\usepackage{caption}   
\usepackage{booktabs}  
\usepackage{appendix}

\title{Jointly Learning to Repair Code and Generate Commit Message}

\author{Jiaqi Bai\textsuperscript{${\dagger\ddagger}$}\thanks{\ \ Contribution during internship at Microsoft Research}, 
Long Zhou\textsuperscript{${\diamondsuit }$}, 
Ambrosio Blanco\textsuperscript{${\diamondsuit }$}, 
Shujie Liu\textsuperscript{${\diamondsuit }$}, \\ 
\textbf{Furu Wei}\textsuperscript{${\diamondsuit }$}, 
\textbf{Ming Zhou}\textsuperscript{${\diamondsuit }$}, \textbf{Zhoujun Li}\textsuperscript{${\dagger\ddagger}$} \\
    \textsuperscript{$\dagger$}School of Cyber Science and Technology, Beihang University, China \\
    \textsuperscript{$\ddagger$}State Key Lab of Software Development Environment, Beihang University, China \\
    \textsuperscript{${\diamondsuit}$}{Microsoft Research Asia} \\

{\tt \{bjq,lizj\}@buaa.edu.cn} \\ 
{\tt \{lozhou,ambrosio.blanco,shujliu,fuwei,mingzhou\}@microsoft.com}
}

\begin{document}
\maketitle
\begin{abstract}
We propose a novel task of jointly repairing program codes and generating commit messages.
Code repair and commit message generation are two essential and related tasks for software development. However, existing work usually performs the two tasks independently.
We construct a multilingual triple dataset including buggy code, fixed code, and commit messages for this novel task.
We provide the cascaded models as baseline, 
which are enhanced with different training approaches,  including the teacher-student method, the multi-task method, and the back-translation method. To deal with the error propagation problem of the cascaded method, the joint model is proposed that can both repair the code and generate the commit message in a unified framework.
Experimental results show that the enhanced cascaded model with teacher-student method and multitask-learning method achieves the best score on different metrics of automated code repair, and the joint model behaves better than the cascaded model on commit message generation.
\end{abstract}

\section{Introduction}

Deep learning has been demonstrated remarkably adept at numerous natural language processing (NLP) tasks, such as machine translation \cite{bahdanau2014neural}, relation extraction \cite{zhang2017end}, grammar error correction \cite{ge2018reaching}, and so on. The success of deep learning in NLP also promotes the development of which in programming languages \cite{DBLP:conf/emnlp/ClementDTSS20,DBLP:journals/corr/abs-2102-04664}.
Recently, researchers have exploited deep learning to programming-language related tasks, such as code completion \cite{svyatkovskiy2020intellicode}, automated code repair \cite{tufano2018empirical}, commit messages generation \cite{xu2019commit}, code search \cite{gu2018deep}, and so on. 
Among these tasks, automated code repair and commit message generation are the two most active and closely related tasks.
The former is to repair software bugs automatically without the intervention of a human programmer. The latter aims to generate natural language descriptions of code changes, which act as a record of feature additions and bug repairs. 

\begin{figure}[t]
\centering
\includegraphics[width=7.5cm]{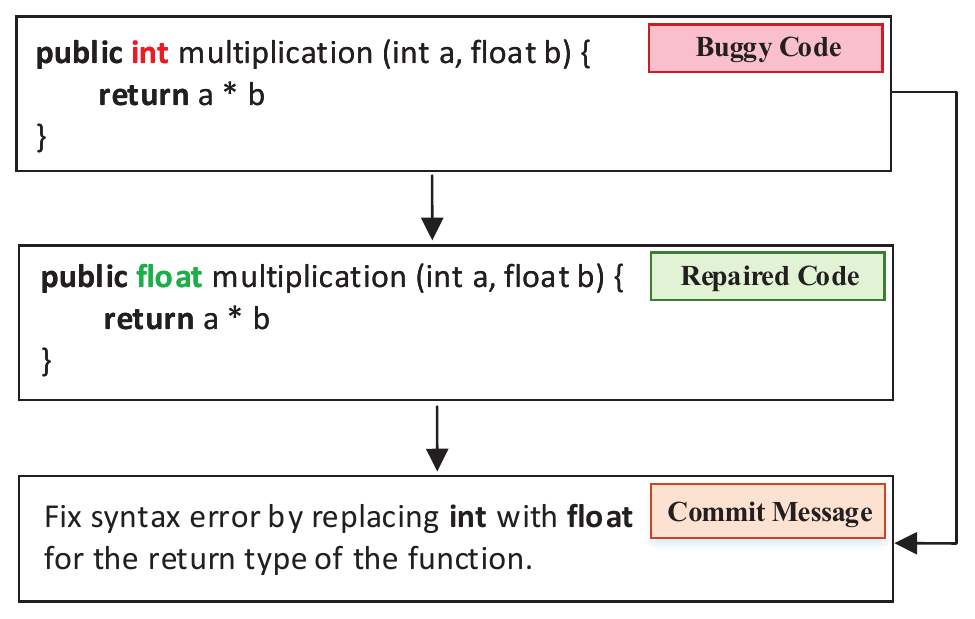}
\caption{An illustrative example for our proposed task. Given a buggy code, the task is to generate its corresponding repaired version, as well as the commit message that describes their changes.}
\label{fig1}
\end{figure}

Because these two tasks can potentially reduce debugging costs in software development and helps programmers to understand the high-level rationale of changes, a lot of great work has been proposed to deal with automated program repair \cite{tufano2018empirical,chen2019sequencer,dinella2020hoppity,yasunaga2020graph,DBLP:conf/acl/TangZBLWZY21} and commit message generation \cite{loyola2017neural,liu2020atom,nie2020contextualized}, respectively.
However, existing work tackles the two tasks independently, ignoring the underlying relationship between these two closely related tasks, e.g., after fixing the bug, commit message can record the process of code repair.
Therefore it is crucial to explore how to bridge these two tasks and achieve the code repair and commit messages generation simultaneously.

In this paper, we formulate a novel task to jointly repair program code and generate commit message, given the buggy code.
To facilitate the study of this task, we create a dataset with multiple programming languages. The dataset is collected from commit and buggy-fixed histories of open-source software projects, where each example consists of buggy code, fixed code, and the corresponding commit message. 
We first introduce the cascaded methods as baseline. The cascaded model employs one model to repair code and the other to generate commit message successively. 
We enhance this cascaded model with three training approaches inspired by the low-resource machine translation, including the teacher-student method \cite{chen2017teacher}, the multi-task learning method \cite{domhan2017using}, and the back-translation method \cite{sennrich2016edinburgh}. To deal with the error propagation problem of the cascaded method, we propose a joint model which can achieve both code repair and commit message generation in a single model. 
We train and evaluate our model using the created triple (\emph{buggy}-\emph{fixed}-\emph{commit}) dataset. The results demonstrate the validity of our proposed methods, which achieve a significant improvement over baseline in both qualities of code and commit messages.
Particularly, the enhanced cascaded method obtains the best performance on code repair task, and the joint method behaves better than the cascaded method on commit message generation task.

Our main contributions are as follows:
\begin{itemize}
\item We propose the novel task of jointly repairing code and generating commit message. Moreover, we collect and release a multilingual \emph{buggy}-\emph{fixed}-\emph{commit} dataset for the task.
\item We perform an empirical study of different machine learning-based methods for code repair and commit message generation.
\item To the best of our knowledge, this is the first work to investigate the effectiveness of joint modeling with the code repair process and commit message generation.
\end{itemize}

\section{Task Definition}
We aim to generate the repaired code and the commit message based on the given buggy code.
For the example in Figure \ref{fig1}, our goal is to replace syntax bug ``int'' with ``float'' for the return type of the ``multiplication'' method, then generate a piece of commit message to describe the change.

Formally, given a triple (\emph{buggy-fixed-commit}) dataset $\mathcal{D}=\left\{ \left( B_i,F_i,C_i \right) \right\} _{i=1}^{K}$, where the $i$-th sample consists of a buggy code snippet $B_i$, its fixed version $F_i$, and a commit message $C_i$ that is used to describe the changes from $B_i$ to $F_i$,
our goal is to learn probability distribution $P\left( \left. C,F \right|B \right)$. 
In practice, the commit message $C$ is hard to estimate without the full consideration of both $B$ and $F$. Therefore, it is a reasonable way to firstly predict the fixed code $F$ based on $B$. Then learn how to generate an appropriate message $C$ according to $B$ and $F$. The probability $P\left( \left. C,F \right|B \right)$ can be decomposed as:
\begin{equation}\small
P(C,F|B)=\sum_F{P\left( \left. F \right|B \right) P\left( \left. C \right|B,F \right)}
\end{equation}
Thus, given a new buggy code $B$, we can generate its fixed version $F$, and the commit message $C$ following the conditional probability $P\left( \left. F \right|B \right)$ and $P\left( \left. C \right|B,F \right)$, respectively.

\section{Approach}
\label{Approach}

In this section, we firstly introduce the cascaded models enhanced by the  teacher-student method, the multi-task learning method, and the back-translation method (Section \ref{cascaded}), which generate repaired code and commit message in a two-stage manner. Then, we propose a joint model (Section \ref{joint}), which is capable of jointly optimizing the generation of repaired code and commit message in an end-to-end manner. The models described in this section are all build on the Transformer model \cite{vaswani2017attention}, where we devise the model to take in some representation of input and then yield a distribution over output vocabulary.

\subsection{Cascaded Model}
\label{cascaded}

\label{chap51}
Cascaded model is one of the most straight-forward methods to tackle this problem, where $F$ is used as a hinge to build bridges between $B$ and $C$. Formally, given the buggy code $B$, the generation of commit message $C$ can be conducted in two steps. The first step aims to generate $F$ conditioned on $B$, which can be defined by minimizing the following negative log-likelihood loss:
\begin{equation}\small
\label{eq2}
\mathcal{L}_{\mathcal{F}}\left( \theta \right) =-\sum_{\left( B,F \right) \in \mathcal{D}}{\log P\left( \left. F \right|B \right)}
\end{equation}

The second step is to generate commit message $C$ based on $B$ and previous generated $F$, and it can be formally expressed as:
\begin{equation}\small
\label{eq3}
\mathcal{L}_{\mathcal{C}}\left( \theta \right) =-\sum_{\left( B,F,C \right) \in \mathcal{D}}{\log P\left( \left. C \right| g(B,F) \right)}
\end{equation}

\noindent where $g(B,F)$ is a function to combine $B$ and $F$ as model input, which could be concatenating them, or using their changing information\footnote{We use \texttt{difflib} to represent code changes. The tool can be found in  \url{https://docs.python.org/3/library/difflib.html}. In this paper, we use the code changes to build the model input, instead of their concatenation, since the latter will result in overlong sequence length, which drops the performance of model by a significant margin.}.
In the following section, the training loss of commit message generation is optimized by Equation (\ref{eq3}), unless explicitly specified.

To further enhance the modeling capabilities of the cascaded model, we introduce three alternative methods by incorporating the teacher-student framework, multi-task learning method, and back-translation method, respectively.

\paragraph{Teacher-student Method}
We attempt to improve the performance of code repair with the help of commit message.
Different from the previous works that directly used comments \cite{guo2020graphcodebert} or compiler error messages \cite{yasunaga2020graph} as the prior information, we utilize the commit message as the posterior information, to supervise the generation of $F$ in code repair. Specifically, the teacher-student framework \cite{hinton2015distilling} is employed to distill knowledge from teacher model to student model, which first learns a teacher model $P\left( \left. F \right|B,C \right)$ with the use of $C$, where $C$ is the truth commit message . Then, the teacher model teaches the student model $P\left( \left. F \right|B \right)$ by minimizing the KL divergence \cite{kullback2006information}, which is defined by
\begin{equation}\small
\mathcal{L}_{KL}\left( \theta \right) =\sum_{\left( B,F,C \right) \in \mathcal{D}}{\begin{array}{c}
	Q\left( \left. F \right|B,C \right) \cdot \log \frac{Q\left( \left. F \right|B,C \right)}{P\left( \left. F \right|B \right)}\\
\end{array}}
\end{equation}

\noindent where $Q\left( \left. F \right|B,C \right)$ represents the teacher's sequence distribution over the sample space of all possible sequences. When optimizing $\mathcal{L}_{KL}$, the posterior distribution $Q\left( \left. F \right|B,C \right)$ can be regarded as labels, so that our model is instructed to use prior distribution $P\left( \left. F \right|B \right)$ to approximate $Q\left( \left. F \right|B,C \right)$ accurately. 
During the training stage of code repair, the student model not only learns from the output probabilities of teacher model, but also learns from the correct context, which is formulated by
\begin{equation}\small
\mathcal{L}_{\mathcal{F}}^{T}\left( \theta \right) =\mathcal{L}_{\mathcal{F}}\left( \theta \right) +\mathcal{L}_{KL}\left( \theta \right) 
\end{equation}

\paragraph{Multi-task Learning}
Inspired by previous work which shows that given the buggy lines can significantly improve the performance of code repair \cite{chen2019sequencer,wen2018context,saha2017elixir}, 
we use an alternative way to improve code repair, which is the multi-task learning method. Specifically, we introduce a line-level binary sequence classification task as an auxiliary learning task to assist code repair, which reduces the difficulties for the model to locate the buggy lines\footnote{To obtain the buggy lines, we employ the \texttt{difflib} to extract the line-level changes from buggy code to its fixed version. We maintain the lines only exist in the buggy version (i.e., remove the lines started with ``+'' and ``?'').}. 
To help the model distinguish from the line-level information and the token-level information, we add the ``\emph{[CLS]}'' token at the beginning of each line of buggy code $B$, which is used to align with the tagging label $T$, where $T\in \left\{ 0,1 \right\}$, in which tag 0 means the line is error-free, and tag 1 means the line is buggy. 
To identify the buggy lines, we build a sequence classifier  based on encoder output to implement the line-level binary sequence tagging task. The line-level sequence classification loss can be defined as:
\begin{equation}\small
\label{mtl}
\mathcal{L}_{\mathcal{T}}\left( \theta \right) =-\sum_{B\in \mathcal{D};T\in \left\{ 0,1 \right\}}{\log P\left( \left. T \right|B \right)}
\end{equation}

At the stage of code repair, we jointly optimize the objective of sequence classification task and sequence generation task, i.e.,
\begin{equation}\small
\mathcal{L}_{\mathcal{F}}^{M}\left( \theta \right) =\mathcal{L}_{\mathcal{F}}\left( \theta \right) +\mathcal{L}_{\mathcal{T}}\left( \theta \right) 
\end{equation}

\paragraph{Back-translation Method}
Back translation has been demonstrated as an effective way on data augmentation \cite{sennrich2016edinburgh,lachaux2020unsupervised}, and it leverages monolingual data to expand as pseudo-parallel data in a weakly-supervised manner. More precisely, we first train a back-directional model, that is a repaired code to buggy code model parameterized by $P\left( \left. B \right|F,\theta _{F\rightarrow B} \right)$. Then, the pseudo-parallel data is created by the back-directional model, in which the repaired code is regarded as the model input, and the goal is to predict its corresponding buggy version, which is formulated by
\begin{equation}\small
\hat{B}=\underset{B}{{\rm{argmax}}}P\left( \left. B \right|F,\theta _{F\rightarrow B} \right) 
\end{equation}

\noindent where $\theta _{F\rightarrow B}$ is the parameter learned by maximum likelihood estimation on $\mathcal{M}$. $\mathcal{M}$ is a non-parallel corpus of fixed code, which is used to build the pseudo-parallel data. After obtaining $\hat{B}$, the pseudo parallel data $\mathcal{P}=\{( \hat{B},F)\}$ is created to merge with the parallel data $\mathcal{D}$ to obtain the augmented parallel data $\mathcal{D}^{\prime}$, which is used to train the code repair model according to Equation \ref{eq2}.

\subsection{Joint Model}
\label{joint}

Although the above three methods can boost the performance of cascaded method, they still suffer from three challenges: (1) the generated fixed code may contain errors, and those errors will be propagated to the next step of commit generation, (2) they lose the inter-dependencies among global features to represent the changing details of code repair during commit message generation,  and (3) the two-stage method results in low decoding efficiency. These problems may lead to the poor performance of commit generation.
To this end, we propose a joint method that incorporates with a novel changes-aware dynamic attention mechanism to jointly decode fixed code and commit message. 

\paragraph{Model Architecture}

\begin{figure}[t]
\centering
\includegraphics[width=7.5cm]{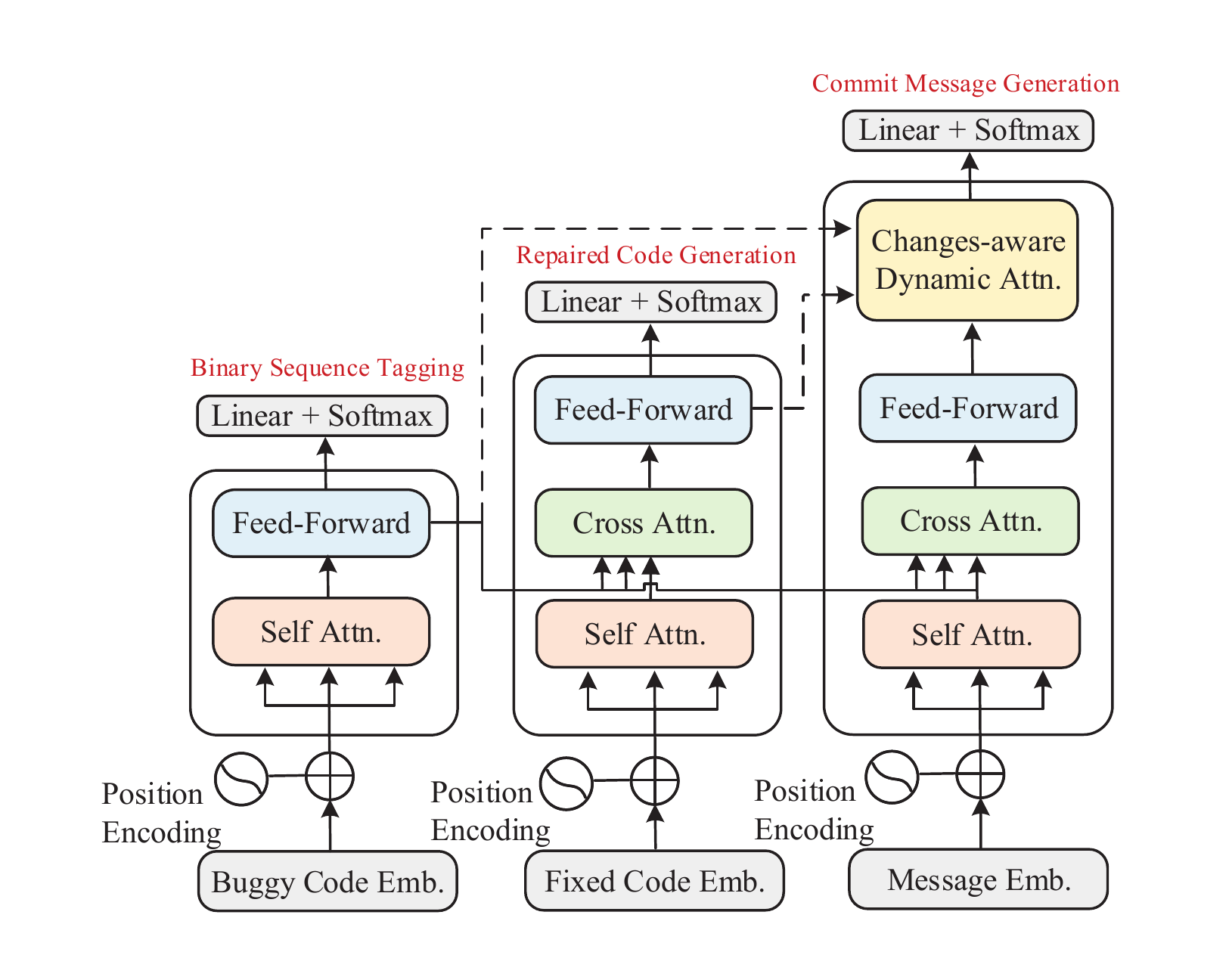}
\caption{The architecture of the proposed joint model, where residual connection and layer normalization are omitted for simplification.}
\label{fig2}
\end{figure}

The overview of our model is shown in Figure \ref{fig2}. Our model consists of three components: a buggy encoder, a fixed decoder, a message decoder with a changes-aware dynamic attention module. 
At first, the buggy encoder is deployed to encode the buggy code $B$, and map it into a sequence of output $\mathbf{z_b}$, where $\mathbf{z}_{\mathbf{b}}\in \mathbb{R}^{n\times H}$, $n$ and $H$ are the length of $B$ and the hidden size of model, respectively. $\mathbf{z_b}$ is used for line-level binary sequence tagging (optimized as Equation \ref{mtl}) and as an indispensable component to produce changes information.
Then, the fixed decoder generates a high-level representation $\mathbf{z_f}$,  $\mathbf{z}_{\mathbf{f}}\in \mathbb{R}^{m\times H}$ is used to generate a repaired code $\hat{F}$, and produce changes information with $\mathbf{z_b}$. 
After that, the commit decoder that combines with the changes-aware dynamic attention mechanism generates an output representation $\mathbf{z_c}$, $\mathbf{z}_{\mathbf{c}}\in \mathbb{R}^{l\times H}$ is used to attend over each representation of $\mathbf{z_b}$ and $\mathbf{z_f}$, then get a final output distribution to generate messages. In the following part, we will introduce our proposed changes-aware dynamic attention mechanism, as well as the method to jointly train our model.

\paragraph{Changes-aware Dynamic Attention}

During decoding the message, the output $\mathbf{z_c}$ generated by the message decoder respectively attends over $\mathbf{z_b}$ and $\mathbf{z_f}$ to obtain the context representation $\mathbf{c_b}$ and $\mathbf{c_f}$ by dot-product attention. which is formulated by
\begin{equation}\small
\mathbf{c}_{\phi}=softmax\left( \frac{\mathbf{z}_c\mathbf{z}_{\phi}^{T}}{\sqrt{H}} \right) \mathbf{z}_{\phi}
\end{equation}

\noindent where $\phi \in \left\{ \mathbf{b},\mathbf{f} \right\} $.  
Similar as \citet{vaswani2017attention}, we use the scaling factor $\sqrt{H}$ to make gradient more stable during optimization. Intuitively, the context vector could provide much information to dynamically indicate the alignments of changes over the attended features during decoding the commit messages. 

We subtract $\mathbf{c}_{\mathbf{b}}$ from $\mathbf{c}_{\mathbf{f}}$ in order to represent the semantic changes that took place from buggy code to its fixed version, and plus $\mathbf{c}_{\mathbf{b}}$ with $\mathbf{c}_{\mathbf{f}}$ to denote the semantic summarization of them, which is defined by
\begin{equation}\small
\begin{array}{c}
\delta =\mathbf{c}_{\mathbf{b}}-\mathbf{c}_{\mathbf{f}}
\\
\zeta =\mathbf{c}_{\mathbf{b}}+\mathbf{c}_{\mathbf{f}}
\end{array}
\end{equation}

\noindent here, the $\delta$ and $\zeta$ are $\mathbb{R}^{l\times H}$ matrices, which represent the changes context and summarization context, respectively. 
Intuitively, the $\delta$ is necessary to generate informative commit messages since it describes the changes from buggy code to its fixed version. Nevertheless, the summarization $\zeta$ is also indispensable during decoding, since it always contain vital information to generate the meaningful tokens (e.g., function name or class name), which may not be included in $\delta$.

We develop the control gates to balance the contribution between $\delta$ and $\zeta$ during decoding the message. The control gates control the degree that attending over each feature of $\delta$ and $\zeta$, which is defined as
\begin{equation}\small
\left[ \begin{array}{c}
	g_{\delta}\\
	g_{\zeta}\\
\end{array} \right] =\sigma \left( \mathbf{W}_g\left[ \delta ;\zeta \right] ^T+\mathbf{b}_g \right) 
\end{equation}

\noindent where $\left[ a;b \right]$ denotes the concatenation between $a$ and $b$. $\mathbf{W}_g\in \mathbb{R}^{2H\times 2H}$ and $\mathbf{b}_g\in \mathbb{R}^{2H\times l}$ are learnable parameters. The gates $g_{\delta}\in \mathbb{R}^{H\times l}$ and $g_{\zeta}\in \mathbb{R}^{H\times l}$ are used to control $\delta$ and $\zeta$, respectively. 

The final output of changes-aware dynamic attention module is the linear combination between the output state of commit message decoder and the gated fusion of context  representations, which can be calculated as:
\begin{equation}\small
\mathbf{o}_{\mathbf{c}}=\mathbf{z}_{\mathbf{c}}+\mathbf{W}_{\mathbf{o}}\left( g_{\delta}\odot \delta ^T+g_{\zeta}\odot \zeta ^T \right) 
\end{equation}
where $\mathbf{W_o}\in \mathbb{R}^{H\times H}$ is the learnable weights. $\odot$ denotes the element-wise product.

\paragraph{Joint Training}
We jointly train our model in an end-to-end manner, the overall loss is defined as
\begin{equation}\small
\label{joint_training_loss}
\mathcal{L}_{\mathcal{J}}\left( \theta \right) =\mathcal{L}_{\mathcal{R}}\left( \theta \right) +\mathcal{L}_{\mathcal{C}}\left( \theta \right) +\mathcal{L}_{\mathcal{T}}\left( \theta \right)
\end{equation}
where $\mathcal{L}_{\mathcal{R}}\left( \theta \right)$, $\mathcal{L}_{\mathcal{C}}\left( \theta \right)$and $\mathcal{L}_{\mathcal{T}}\left( \theta \right)$ are used to optimize the repaired code generation, commit message generation, and binary sequence classification, respectively. 
When training multilingual model of fixing code and predicting commit message, following multilingual neural machine translation \cite{Johnson2017GooglesMN}, we mix the training corpus and add a special token (e.g., \texttt{<java>}) at the beginning of each input sequence to distinguish from different programming languages.

\section{Data}
\label{sec_Data}
In this section, we describe the creation of the dataset in detail. We first describe how we collect the data in the wild. Then, we introduce the preparation process of the data to make it suitable for our tasks.

\paragraph{Data Collection}
We collected data from GitHub Archive\footnote{\url{https://www.githubarchive.org}} using the GH Archive API.
We first download the event data in each public event from 2018 to 2020, and then filter them using the Github API\footnote{\url{https://docs.github.com/en/free-pro-team@latest/rest}} to obtain meta information of each commit.
Specifically, we maintain the commits that consist of edits to the files with multiple programming languages. (i.e., Java, Javascript, Python, C sharp, Cpp). 
Moreover, to ensure that the prior file and post file are repairing code, we follow \cite{fischer2003populating}, where the commits without the specific patterns (i.e., “fix” or “solve”) in its commit messages are filter out.
After we obtain the meta information of the filtered commit, we begin downloading the buggy file (i.e., the file prior to the commit) and fixed file (i.e., the file following the commit) in pair.
Apart from the above multilingual dataset, we also build a Java-only monolingual triple dataset from the corpus (buggy-fixed pair) released by \citet{tufano2018empirical}\footnote{\url{https://sites.google.com/view/learning-fixes/data}}.

\paragraph{Data Preparation}

\begin{table}[t]\small
\centering
\begin{tabular}{cccccc}
    \toprule
    ~ & Languages & Train & Valid & Test & Total \\
    \midrule
    \multirow{5}*{Multi.} & Python & 36682 & 4585 & 4586 & 45853 \\
    ~ & Java & 11129 & 1391 & 1392 & 13912 \\
    ~ & Javascript & 21446 & 2680 & 2681 & 26807 \\
    ~ & C-sharp & 5424 & 678 & 678 & 6780 \\
    ~ & Cpp & 8510 & 1063 & 1064 & 10637 \\
    \midrule
    Mono. & Java & 47775 & 3000 & 3000 & 53775 \\
    \bottomrule
\end{tabular}
\caption{Data statistic of the multilingual and the monolingual dataset.}
\label{data_statistic}
\end{table}
The commit messages are filtered by (i) removing the messages whose length shorter than 3 words or longer than 100 words; (ii) filtering the \texttt{url} in commit messages; (iii) removing the messages that appear more than 3 times in the dataset. The rationale behind the latter decision was to remove the data with meaningless commit messages (e.g., ``fix bug.'', ``fix an error.'', etc.).
For the processing of file-level buggy code and fixed code, we follow \cite{tufano2018empirical}, where both the buggy code and fixed code are separated into method-level fragments since the file-level granularity is too large to learn patterns of transformation. 
After preparation, we obtain the clean triples consist of buggy code, fixed code, and commit message. The statistics of the dataset used in this paper are summarized in Table \ref{data_statistic}.
More processing details and statistics can be found in Appendix \ref{appdx_dataset_details} and Appendix \ref{appdx_dataset_statistics}.
We release the datasets at \url{https://github.com/jqbemnlp/BFCsData}.

\section{Experiments}

\begin{table*}[t]\small
\centering
\renewcommand\arraystretch{1.2}
\begin{tabular}{|l|cc|cc|}
    \hline

    \multirow{3}*{Models} & 
    \multicolumn{2}{c}{Automated Code Repair} &
    \multicolumn{2}{|c|}{Commit Message Generation} \\ 
    \cline{2-3} \cline{4-5}
    ~ & BLEU-4 & xMatch & BLEU-4 & ROUGE-L \\

    \hline
    
    Naive Method & 87.45 & 0.00 \
    & 8.40 & 7.98 \\
    
    Oracle Method& - & - \
    & 12.64 & 11.59 \\
    
    \hline

    Cascaded Model & 85.07 & 3.21 \
    & 9.69 & 9.41 \\

    \quad + Teacher-student & \textbf{88.23} & 6.16 \
    & 10.58 & 10.19 \\

    \quad + Multitask & 87.94 & \textbf{8.33} \
    & 10.36 & 10.1 \\
    
    \quad + Back-translation & 87.73 & 5.26 \
    & 10.19 & 9.84 \\
    
    \hline
    
    Joint Model & 87.61 & 8.01 \
    & \textbf{11.48}* & \textbf{10.62}* \\

    \hline
\end{tabular}
\caption{Results of the cascaded model and the proposed joint model
on the monolingual dataset for code repair and commit message generation tasks. The bold face indicates the best result under the corresponding metric. Significant improvements over the best baseline results are marked with * (t-test, p<0.05).
}
\label{table2}
\end{table*}

\subsection{Experimental Settings}

\paragraph{Evaluation Metrics}
We conduct evaluations on both code repair and commit message generation. For the code repair, we use exact match accuracy \cite{chen2018tree} to measure the percentage of the predicted fixed code that are exactly matching the truth fixed code. In addition, we also introduce the BLEU-4 score \cite{papineni2002bleu} as a supplementary metric to evaluate their partial match.
For the commit message generation, we use BLEU-4 and Rouge-L \cite{lin2004rouge} to evaluate our model.

\paragraph{Implementation Details}
All models are implemented using Pytorch framework\footnote{An open-source deep learning platform (\url{https://pytorch.org/})}, trained on four GPUs of NVIDIA Tesla V100.
We use Byte Pair Encoding (BPE)\footnote{The BPE codes are learned by fastBPE (\url{https://github.com/glample/fastBPE}).} \cite{sennrich2016neural} to encode input using a shared vocabulary with 50K symbols. 
The Transformer structure and the hyperparameters are following the default setting in the open-source implementation of XLM\footnote{\url{https://github.com/facebookresearch/XLM}} \cite{lample2019cross}, apart from the embedding dimension and maximum sequence length, which are set as 256 and 512, respectively. 
More training details can be found in Appendix \ref{appdx_hyper_setting}.

\begin{table*}[t]\small
\centering
\renewcommand\arraystretch{1.2}
\begin{tabular}{|l|ccc|ccc|ccc|ccc|}
    \hline
    
    \multirow{4}*{Langs.} \
    & \multicolumn{6}{c}{Automated Code Repair} & \multicolumn{6}{|c|}{Commit Message Generation} \\
    
    \cline{2-7} \cline{8-13} 
    
    & \multicolumn{3}{c}{BLEU-4} \    
    & \multicolumn{3}{|c}{xMatch} \   
    & \multicolumn{3}{|c}{BLEU-4} \   
    & \multicolumn{3}{|c|}{ROUGE-L} \\ 

    ~ & mono. & multi. & $\Delta$ \   
    & mono. & multi. & $\Delta$ \     
    & mono. & multi. & $\Delta$ \     
    & mono. & multi. & $\Delta$ \\    

    \cline{1-1}
    \cline{2-4} 
    \cline{5-7} 
    \cline{8-10}
    \cline{11-13}

    python & 95.21 & 94.99  & -0.22 \
    & 8.32 & 8.01 & -0.31 \
    & 13.29 & 14.01  & 0.72 \
    & 12.83 & 13.46 & 0.63 \\
    
    javascript & 94.89 & 95.21  & 0.32 \
    & 6.78 & 7.42 & 0.64 \
    & 11.03 & 11.63  & 0.60 \
    & 10.79 & 11.30 & 0.51 \\
    
    java & 95.72 & 96.74  & 1.02 \
    & 6.33 & 7.82 & 1.49 \
    & 12.26 & 13.79  & 1.53 \
    & 11.72 & 12.73 & 1.01 \\
    
    cpp & 94.10 & 95.45  & 1.35 \
    & 5.63 & 7.34 & 1.71 \
    & 9.71 & 11.04  & 1.33 \
    & 8.63 & 9.84 & 1.21 \\
    
    c-sharp & 93.26 & 95.34  & 2.08 \
    & 3.98 & 6.92 & 2.94 \
    & 8.13 & 10.98  & 2.85 \
    & 7.19 & 9.93 & 2.74 \\
    
    \hline
\end{tabular}
\caption{Results on the multilingual dataset for both code repair and commit message generation. }
\label{multi_results}
\end{table*}

\subsection{Results on Monolingual Dataset}

We first compare the performance of our proposed cascaded model and joint model for code repair and commit message generation tasks on the monolingual dataset, as listed in Table \ref{table2}.

\paragraph{Automated Code Repair} 
The teacher-student model achieves the highest score on the metric BLEU-4, which indicates that the commit message could provide effective guidance for the model to repair code. Moreover, it also indicates that the teacher-student method successfully distills the knowledge from the teacher model to the student model, without much loss of accuracy\footnote{We have evaluated the performance of both the teacher model and student model, the results show that the student model obtains the comparable performance with teacher model.}. 
The multi-task learning model outperforms other experimental models on metric exact match, and get the comparable performance of the teacher-student model on BLEU-4. The intuition behind that is the model has learned the location information of buggy line from the supervision of line-level binary sequence classification task, which could provide potential guidance for the model to correctly repair code. It is worth noting that the Naive method, which directly uses the buggy code to compare with its repaired version, also gets the distinct score on metric BLEU-4, which indicates the high overlap ratio between the buggy-fixed pairs. 

\paragraph{Commit Message Generation}
Both the joint model and cascaded model are superior to generate meaningful commit message than the naive method, which directly uses buggy code to realize message generation.
The joint model outperforms cascaded models over all evaluation metrics of commit message generation. Specifically, it achieves about 10.8\% and 5.1\% improvements respectively on BLEU-4 and ROUGE-L compared to the multitask learning model, which is one of the most competitive models on code repair. It is highly likely that the joint model effectively captures the inter-dependencies among global features to represent the changing details of code repair during commit message generation, thereof mitigates the error propagation of the two-stage model.

\subsection{Results on Multilingual Dataset}
We further analyze the results of the joint model on the multilingual dataset for both code repair and commit message generation. Table \ref{multi_results} shows the results on five program languages.

With regard to code repair, the result shows that the multilingual model
achieves significant improvements compared to the monolingual model
in terms of java, cpp and c-sharp dataset, and obtains comparable performance on python and javascript dataset, whether using BLEU-4 or exact match as evaluation metric. 
The intuition behind that is the corpus mixed with the multiple programming languages is helpful to make up for the lack of monolingual data during repairing code. In other words, the model could learn the potential bug-fixing patterns from multiple languages, and apply them to the limited monolingual data to handle the deficiency of data-limitation problem. A similar observation can also be found during generating commit messages. As shown in Table \ref{multi_results}, for commit message generation task, the multilingual model outperforms monolingual model over all evaluation metrics and languages. 
We believe that the alignment of embedding spaces across multiple programming languages, shares either the same alphabet or anchor tokens such as variable, digits, string, method name, etc., which allows the model to learn these alignments simultaneously during generating commit messages.

\subsection{Discussion}

\paragraph{Ablation Study}

To further study the effects brought by
different techniques, we show in Table \ref{ablation_study} the result
of different joint model variants on the monolingual dataset.
First, we remove the line-level sequence classification task from our joint model. Therefore, the model is optimized without the loss function $\mathcal{L}_{\mathcal{T}}\left( \theta \right)$ that is mentioned in Equation \ref{joint_training_loss}. We observe that the results of both code repair and commit message generation decrease distinctly,
which shows that locating the buggy lines is important for repairing code and generating commit messages.
Then, we remove the proposed changes-aware dynamic attention module. It can be seen that this modification doesn't impact too much for the code repair, but affect the performance of commit message generation by a large margin. The main reason is that the changes-aware dynamic attention module could effectively model the changes from buggy code to its fixed version,  thereby improves the performance of commit message generation.

\begin{table}[t]\small
\centering
\renewcommand\arraystretch{1.2}
\setlength{\tabcolsep}{1.0pt}
\resizebox{1.0\linewidth}{!}{
\begin{tabular}{|l|cc|cc|}
    \hline

    \multirow{3}*{Models} & \multicolumn{2}{c}{Code Repair} \
    & \multicolumn{2}{|c|}{Message Generation} \\ 
    \cline{2-3} \cline{4-5}
    ~ & BLEU-4 & xMatch \
    & BLEU-4 & ROUGE-L \\

    \hline

    Joint Model & 87.61 & \textbf{8.01} \
    & \textbf{11.48} & \textbf{10.62} \\

    \quad - Binary Tagging & 85.64 & 3.98 \
    & 10.10 & 9.25 \\

    \quad - Changes-aware Attn & \textbf{87.88} & 7.94 \
    & 9.03 & 8.67 \\

    \hline
\end{tabular}
}
\caption{Ablation Study for joint model on monolingual dataset. ``-'' means remove the corresponding part separately.}
\label{ablation_study}
\vspace{-4mm}
\end{table}

\paragraph{Lengths Analysis}

\begin{figure}
\centering
    \subfigure[Code Repair]
    {
    \begin{minipage}{3.6cm}
    \label{len_study_sub1}
    \centering
    \includegraphics[width=3.6cm]{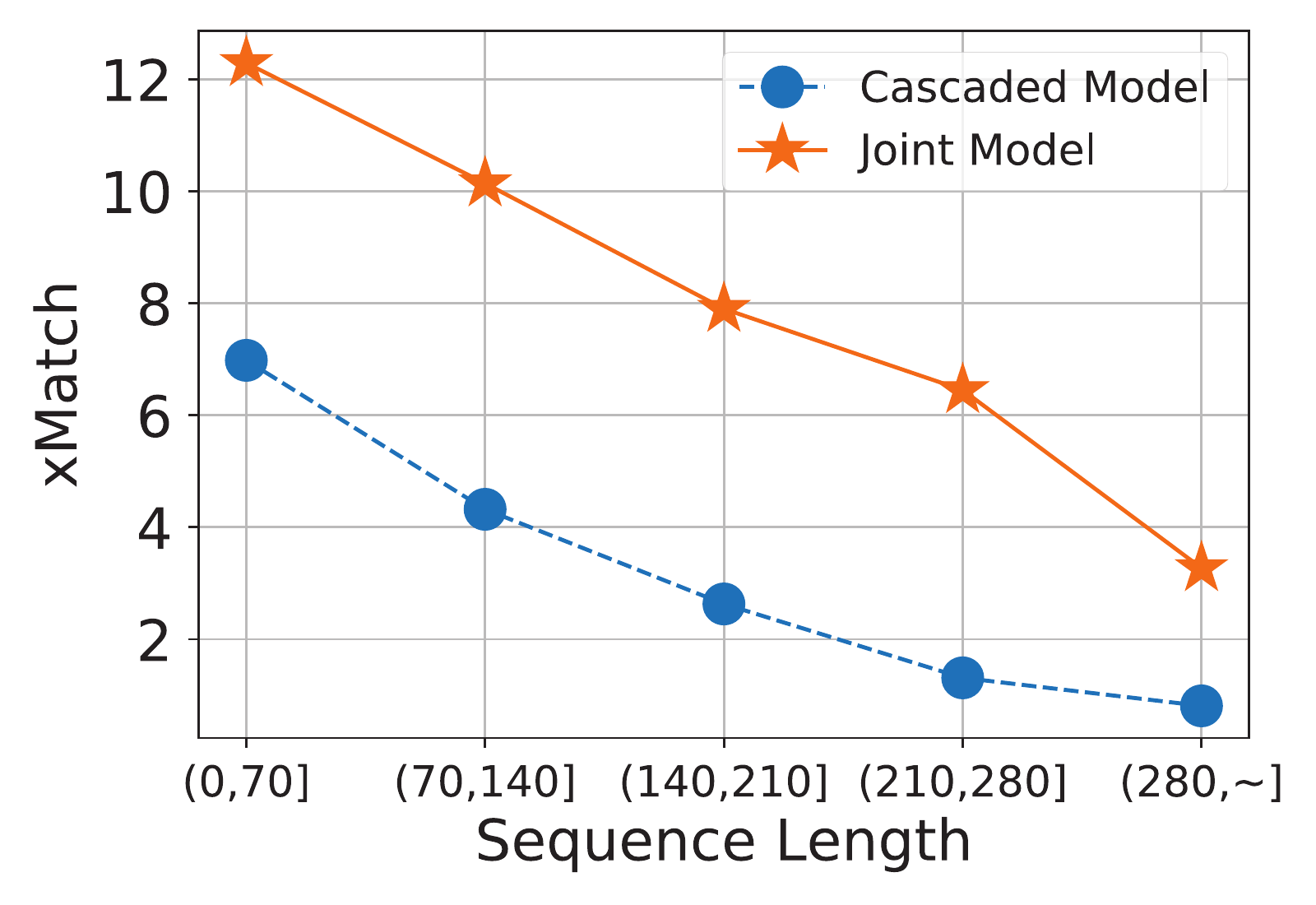}
    \end{minipage}
    }
    \subfigure[Message Generation]
    {
    \begin{minipage}{3.6cm}
    \label{len_study_sub2}
    \centering
    \includegraphics[width=3.6cm]{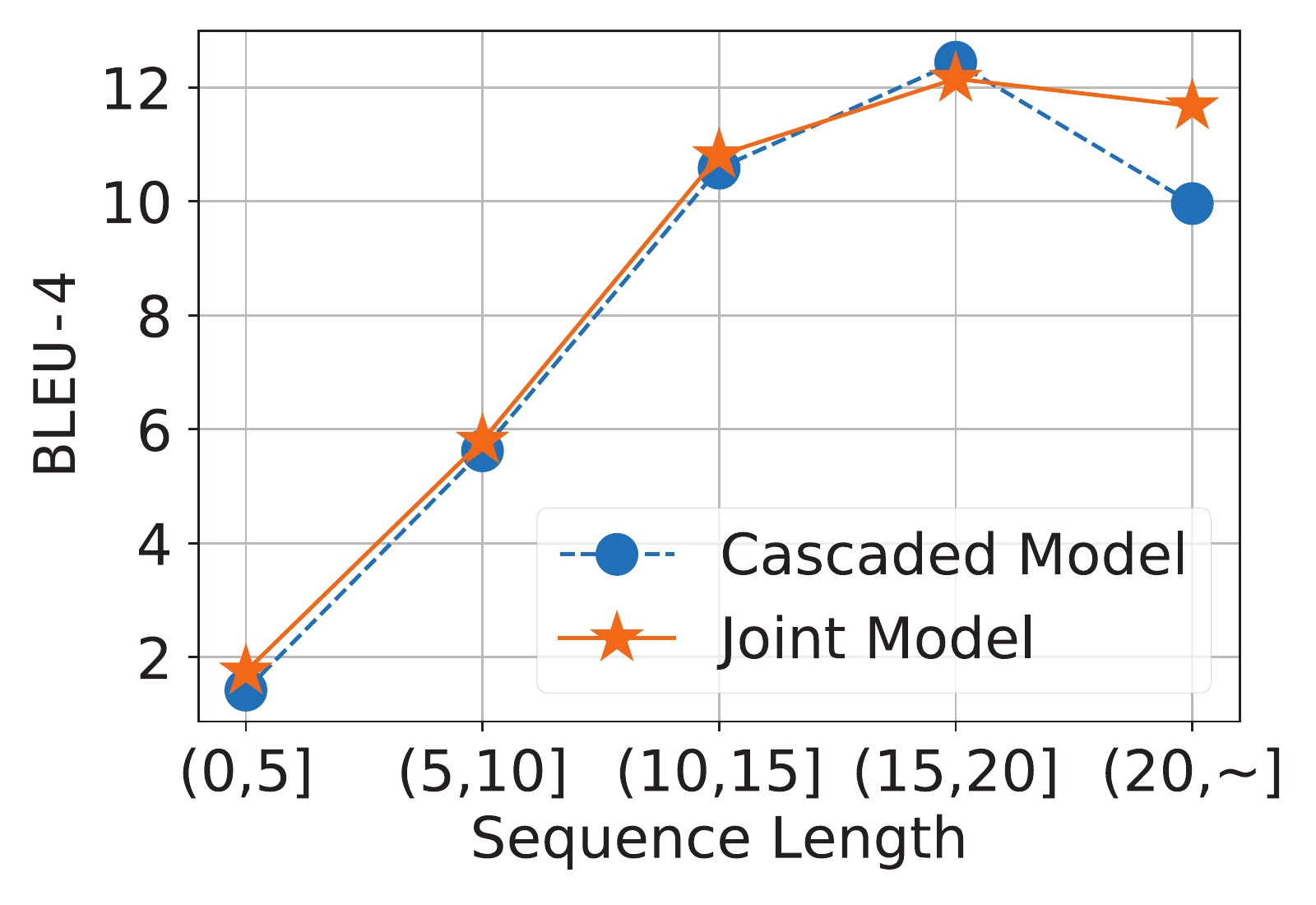}
    \end{minipage}
    }
\caption{Length studies for both code repair and commit message generation.}
\label{length_study}
\vspace{-3mm}
\end{figure}

We further analyze the model performance on different sequence length, and conduct a comprehensive study for both code repair and commit message generation on the monolingual dataset. Figure \ref{len_study_sub1} and \ref{len_study_sub2} present the results of code repair and commit message generation, respectively. Figure \ref{len_study_sub1} demonstrates the challenge of this task, especially when the repaired code with a long sequence length. It can be seen that even the exact match score of the two models declined with the growing length of repaired code, the joint model still outperforms the cascaded model over all length ranges, which demonstrates the stronger capability of the joint model on modeling the code repair.
Figure \ref{len_study_sub2} presents the comparative results of the cascaded model and joint model on generating commit message. We observe that the joint model outperforms the cascaded model when the sequence length exceeds 20, which demonstrates the superiority of the joint model to excavate the underlying semantic relationships between code changes and their corresponding commit message during handling with the long message generation.

\paragraph{Case Study}

\begin{figure}[t]
\centering
\includegraphics[width=7cm]{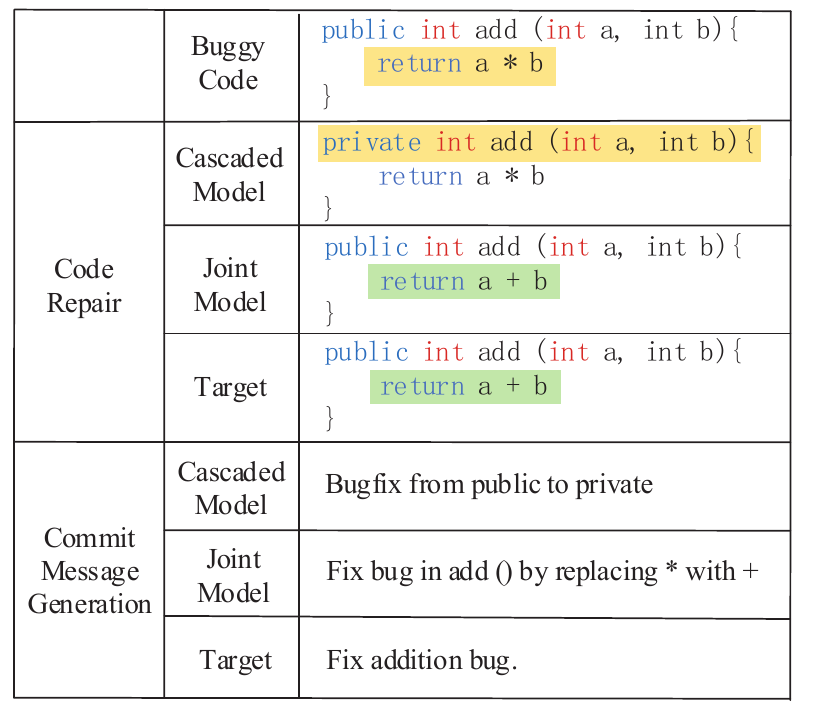}
\caption{Examples on the monolingual dataset. The repaired lines are highlighted with different colors, where yellow means the buggy code is wrongly repaired, while green means it was correctly repaired.}
\label{fig_case1}
\vspace{-3mm}
\end{figure}

We conduct case study on both monolingual and multilingual dataset. Figure \ref{fig_case1} presents the results on monolingual datasets. With regard to the code repair part, the cascaded model wrongly modify the code by replacing \texttt{public} with \texttt{private} in the first line of buggy code, which indicates that it is hard for the model to locate and repair the buggy code without giving any prior information. It is worth noting that the joint model correctly repairs code, 
we believe that the line-level binary sequence classification task assists the model in locating the buggy lines, thereby improving the model's performance during repair the code. As for commit message generation, the joint model successfully captures the changes from buggy code to its fixed version and generates an appropriate message, while the cascaded model fails may due to the error propagation.
More examples on multilingual dataset are shown in Appendix \ref{appdx_case_study}.

\section{Related Work}

Our work is enlightened from two research lines of studies, which are automated code repair and commit message generation. We discuss these topics in the following.

\paragraph{Automated Code Repair}
Conventional approaches mainly focused on a relatively limited and manually-craft set of fixing patterns, which can only fix bugs in a given language or a specific application domain \cite{saha2017elixir,jin2011automated,nguyen2013semfix}.
Very recently, deep learning based approaches are proposed to automatically repair code by learning from massive open-source projects with numerous buggy-fixes pairs \cite{tufano2018empirical,chen2019sequencer,vasic2019neural,yasunaga2020graph}. 
\citet{tufano2018empirical} first proposed using end-to-end neural machine translation model for learning bug-fixing patches.
Besides, \citet{guo2020graphcodebert} demonstrated that appropriately incorporating the natural language descriptions into the pre-train model could further improve the performance of code repair.
\paragraph{Commit Message Generation}
Early work on automatic commit message generation translates source code changes (such as feature additions and bug repairs) into natural language based on pre-defined rules and
templates \cite{buse2010automatically,cortes2014automatically}.
To overcome the limitation of high complexity and difficult extensibility, some researchers employ information retrieval methods to generate commit messages, which attempts to re-use the commit messages of similar code changes \cite{huang2017mining}.
Recent work has focused on adopting machine learning based techniques for the commit message generation problem, which
usually train a sequence-to-sequence model to translate the source changes into commit messages \cite{jiang2017automatically,loyola2017neural,xu2019commit}.

Although automated code repair and commit message generation have achieved rapid development in recent years, existing work usually regards them as two separate tasks and ignores the potential relationship between them.
Different from previous work, we attempt to bridge the two tasks since commit message can be used to record the process of code repair.
Specifically, we propose a novel task to repair code and generate commit message simultaneously with the proposed cascaded and joint methods, based on our collected \emph{buggy}-\emph{fixed}-\emph{commit} dataset.

\section{Conclusion}
In this paper, we propose a novel task to jointly repair code and generate commit message. 
We provide several competitive architectures, including cascaded model and joint model. 
To train and evaluate our models, we collect a multilingual \emph{buggy}-\emph{fixed}-\emph{commit} dataset from Github. The empirical study is conducted to demonstrate the effectiveness of our proposed methods. 
For future work, we plan to incorporate the tree structure of code into the task and employ more indicative metrics \cite{ren2020codebleu} to evaluate the model performance.

\section{Acknowledgement}
This work was supported in part by the National Natural Science Foundation of China (Grant Nos.U1636211, 61672081,61370126), the 2020 Tencent Wechat Rhino-Bird Focused Research Program, and the Fund of the State Key Laboratory of Software Development Environment (Grant No. SKLSDE-2021ZX-18). 

\bibliography{anthology}
\bibliographystyle{acl_natbib}

\clearpage

\begin{appendices}

\section{Data Processing Details}
\label{appdx_dataset_details}

To ensure the commit message describes the changes information that took place from buggy method to its fixed version, we only consider the changes that are inside of a single method in the file. 
Changes that involve multiple methods are not considered in our work, since it is implicit to indicate which method does the commit message describe to. 
Besides, we also develop a heuristics method in which the lexical overlap is employed to filter the commits that the commit message doesn't describe the changes from the buggy method to its fixed version. 
Specifically, we first tokenize the commit message and the code by \texttt{nltk}\footnote{\url{https://www.nltk.org/}} and  \texttt{pygments}\footnote{\url{https://pygments.org/}}, respectively. 
Then, we only maintain the commit in which at least one of the tokens in the commit message matches a code token belonging to the buggy code or repaired code\footnote{During filtering commits, we have removed meaningless tokens in a  commit message, such as punctuation, stop words, url, changes ID, etc., which avoids meaningless tokens affect the quality of filtered results}.
In order to check whether the message describes the changes from buggy code to its fixed version, we randomly selected 100 samples and employ two well-educated annotators for independently analyzing the identified commits. After solving 4 cases of disagreement, they concluded that $97 \%$ of the identified commits were true positive. 

\section{Data Statistics}
\label{appdx_dataset_statistics}

\begin{table}[h]\small 
\centering
\renewcommand\arraystretch{1.3}
\setlength{\tabcolsep}{1.2pt}
\resizebox{1.0\linewidth}{!}{
\begin{tabular}{|l|ccccc|c|}
    \hline

    \multirow{3}*{Data Statistics} & 
    \multicolumn{5}{c|}{Multi.} & 
    \multirow{2}*{Mono.} \\
    
    \cline{2-6} 
    ~ & py & js & java & cpp & c-sharp & ~ \\

    \hline
    
    avg. \# tokens per buggy & 144.7 & 149.8 & 135.0 & 153.5 & 140.7 & 166.9 \\
    
    avg. \# LOC per buggy & 16.9 & 19.7 & 15.7 & 17.8 & 18.5 & 12.6 \\
    
    avg. \# tokens per commit & 12.0 & 9.5 & 12.6 & 15.6 & 10.6 & 13.8 \\
    
    \hline
\end{tabular}
}
\caption{Overview of the multilingual and monolingual datasets. ``LOC'' denotes the physical lines of code.}
\label{avg_statistics}
\end{table}

\begin{figure}[t]
\centering
\includegraphics[width=7cm]{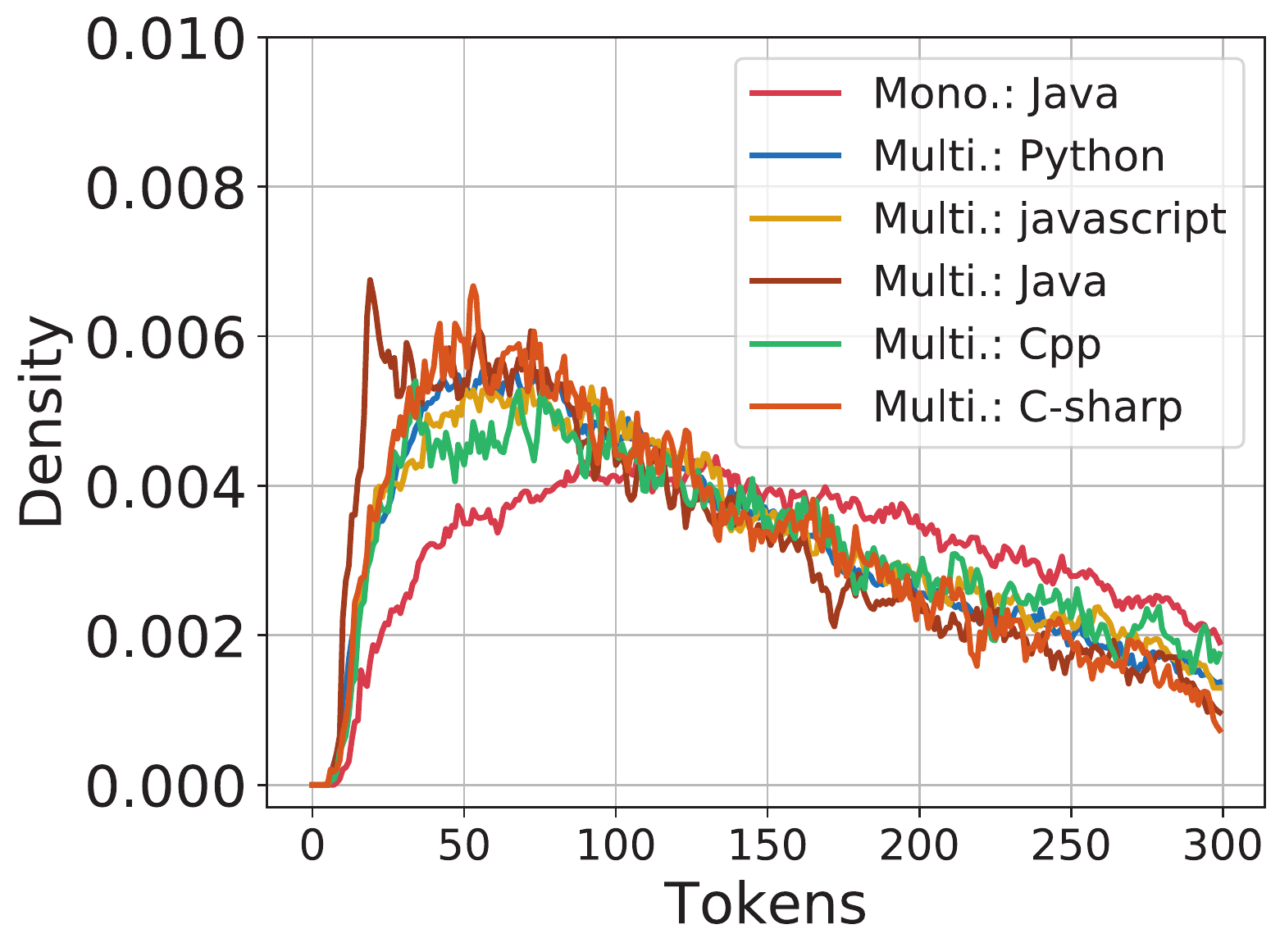}
\caption{The smoothed distribution of buggy code in terms of their size. The $x$-axis denotes the number of tokens for the buggy code. The $y$-axis indicates the density of the buggy code with the corresponding amount of tokens.} 
\label{buggy_distri_plot}
\end{figure}

\begin{figure}[t]
\centering
\includegraphics[width=7cm]{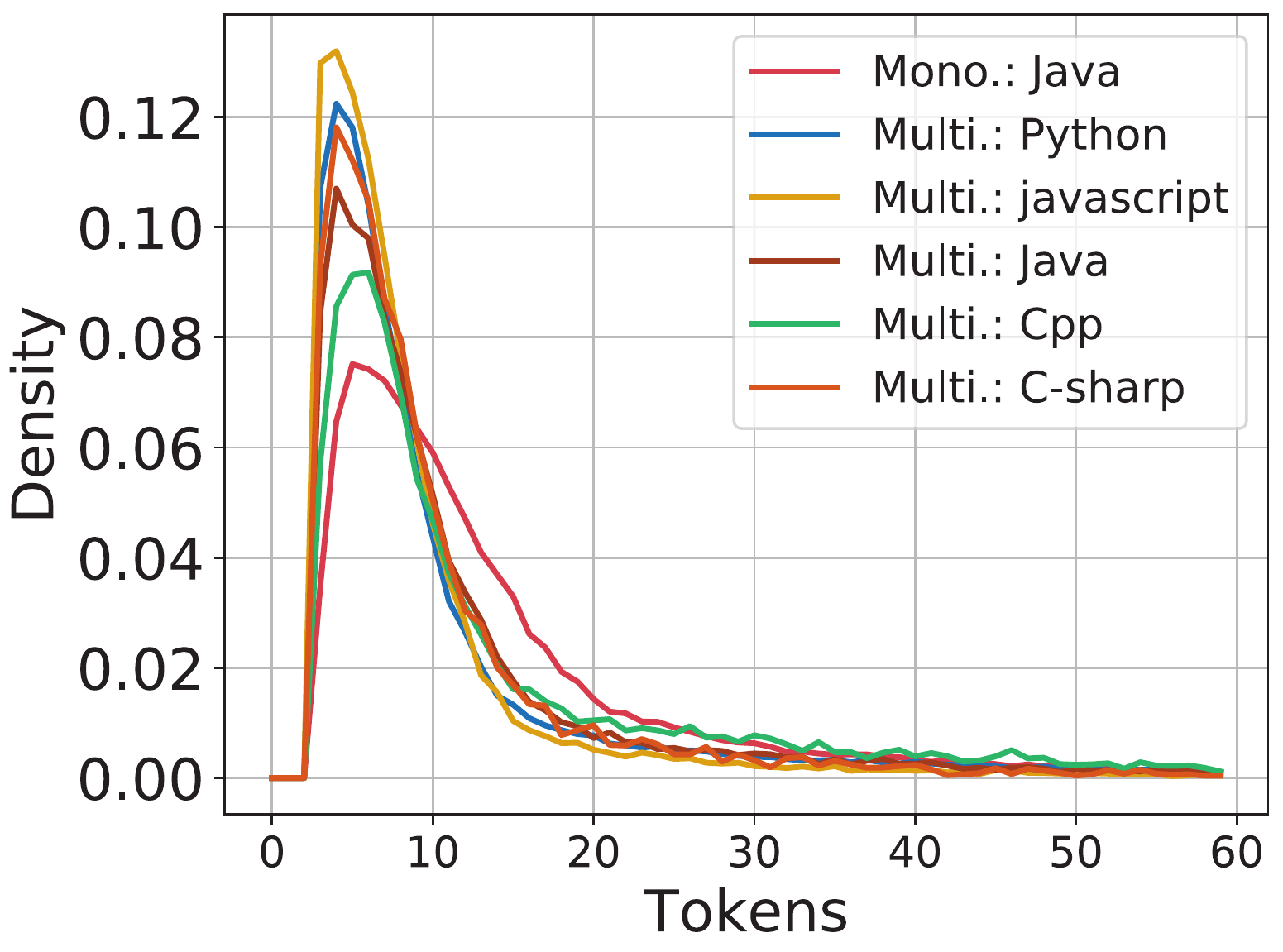}
\caption{The distribution of commit message based on their size.} 
\label{commit_distri_plot}
\end{figure}

\begin{figure}[t]
\centering
\includegraphics[width=7cm]{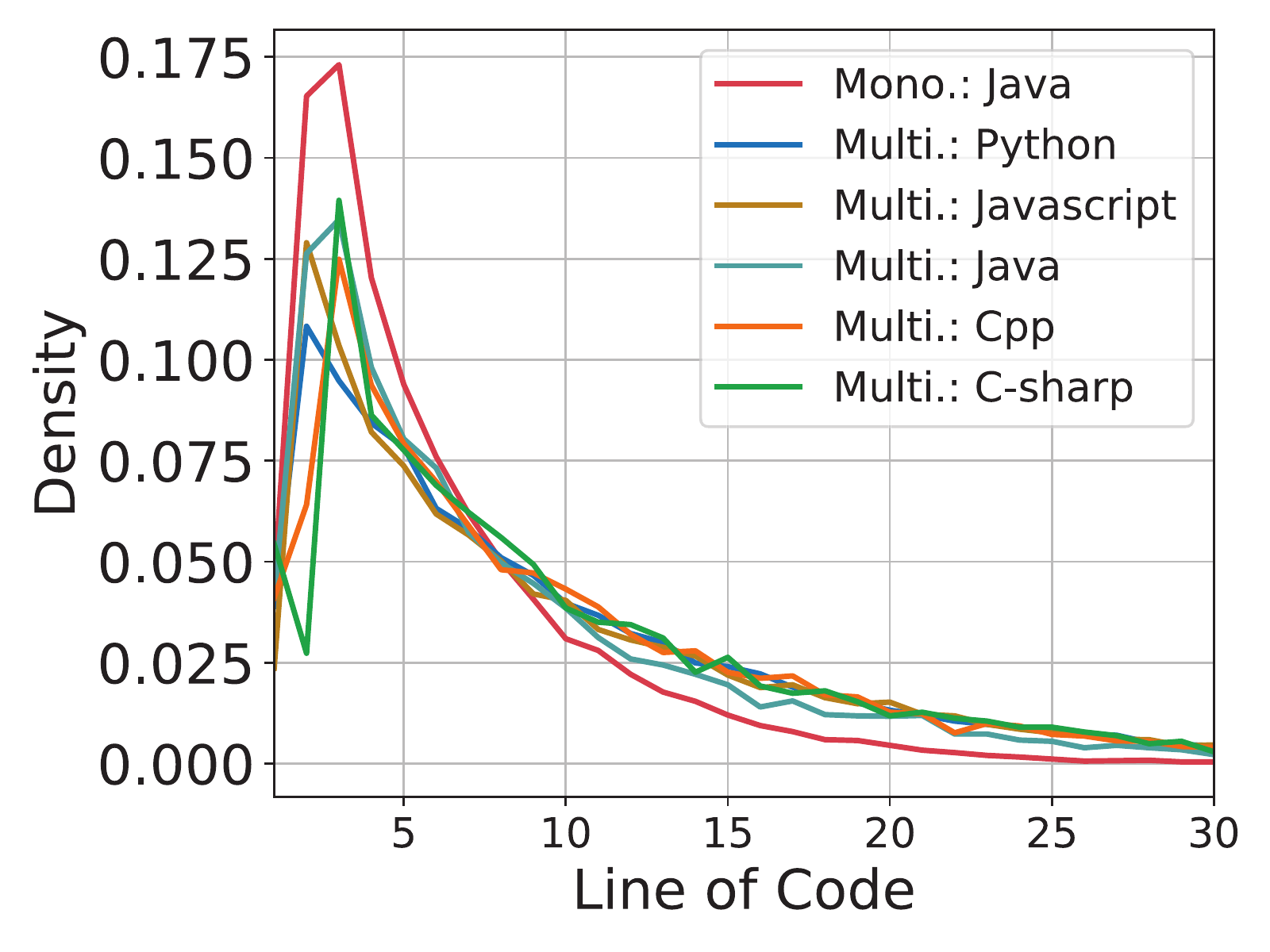}
\caption{The distribution of the line of code that is first appeared in the buggy.} 
\label{buggy_first_loc}
\end{figure}

We take the further analysis for the monolingual and multilingual datasets. Table \ref{avg_statistics} summarizes the average number of tokens per buggy, the average line of code per buggy, and the average number of tokens per commit.
We observe that the average number of lines and tokens for buggy code are considerable, which indicates the difficulty of this task.
Figure \ref{buggy_distri_plot} and Figure \ref{commit_distri_plot} present the distribution of the amount of token for buggy code and commit message, respectively. 
We observe that the density for the buggy code has a long tail that extends over 300 tokens, while the density for the commit message has a peak before 10 tokens.
Figure \ref{buggy_first_loc} shows the distribution of the line number for the first appeared buggy line. It can be seen that the density for the line number also has a peak before the fifth line of buggy code.

\section{Hyperparameter Settings}
\label{appdx_hyper_setting}
\begin{figure}[t]
\centering
\includegraphics[width=7cm]{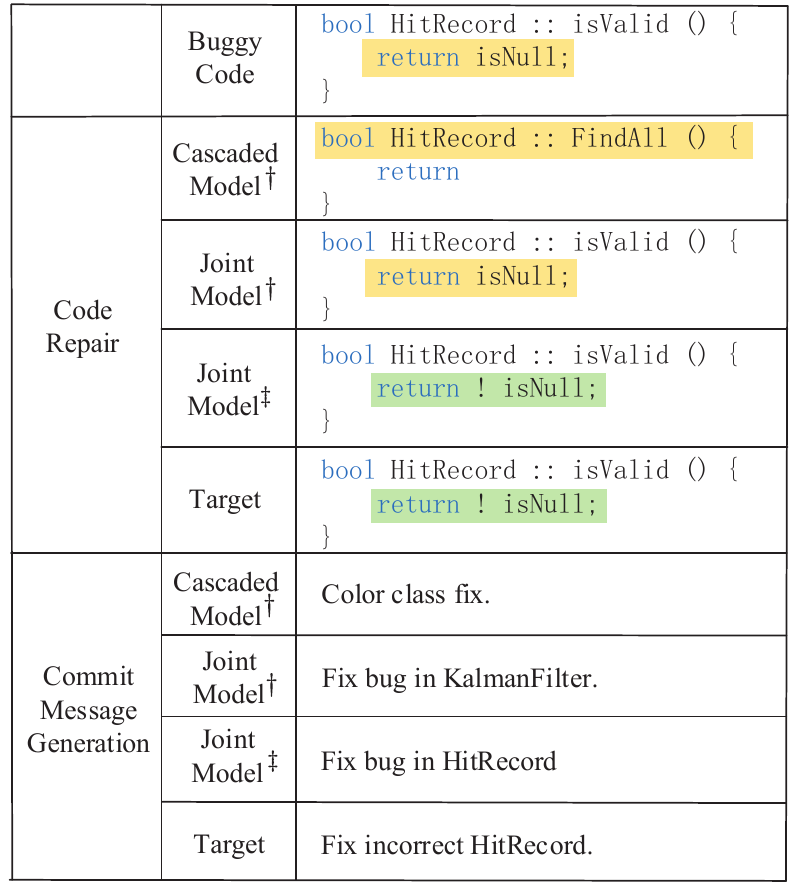}
\caption{Examples on our crawled Cpp dataset. $\dagger$ and $\ddagger$ mean the model is running under the monolingual setting and multilingual settings, respectively.}
\label{fig_case2}
\end{figure}

We train our models\footnote{The base Transformer model has about 40M parameters and the joint model introduce about 0.8M parameters over it.} using Adam optimizer, the initial learning rate is $3\times 10^{-4}$. The mini-batch size and the dropout rate are 16 and 0.1, respectively. We train our models for a maximum of 50 epochs\footnote{It takes about 30 minutes to train an epoch.}. To avoid overfitting, we implement the early stop if the validation performance does not increase for 10 consecutive iterations. 

\section{Case Study}
\label{appdx_case_study}

To further analyze our model under a low-resource setting, we present an example collected from the Cpp dataset. As shown in Figure \ref{fig_case2}, both the pipeline-based model and joint model, which are under the monolingual setting, fail to correctly repair code and appropriately generate commit messages, the most likely reason is that the lacking amount of data doesn't allow the model to successfully capture the useful patterns which are adapted to the specific task. Expectantly, The joint model under a multilingual setting successfully solves the returned value bug. Besides, it captures the key words ``\textbf{HitRecord}'' in generated commit message, which makes the message more relative to the code context.
The example demonstrates that the model performs better
under a multilingual setting compared to which under the monolingual setting, especially on the condition that the amount of monolingual data is limited.

\end{appendices}
\end{document}